%% file: main.tex
\documentclass{article}

\usepackage{spconf}
\usepackage{multirow}
\usepackage{booktabs}
\usepackage{xcolor}
\usepackage{float}
\usepackage{amsmath}
\usepackage{amsfonts}
\usepackage{amssymb}
\usepackage{graphicx}

\newcommand{\inlayer}{\textsc{In}~}
\newcommand{\midlayer}{\textsc{Mid}~}
\newcommand{\outlayer}{\textsc{Out}~}

\input{definitions}

\title{To Reverse the Gradient or Not:
An Empirical Comparison of Adversarial and Multi-Task Learning in Speech Recognition}

\name{*Yossi Adi$^{1,2}$\thanks{Work conducted while Yossi Adi was an Intern at Facebook AI Research.}, *Neil Zeghidour$^{2,3}$~\thanks{*Equal contribution}, Ronan Collobert$^2$, Nicolas Usunier$^2$, Vitaliy Liptchinsky$^2$, Gabriel Synnaeve$^2$}
\address{$^1$Bar-Ilan University. \\$^2$Facebook AI Research. \\$^3$CoML, ENS/INRIA.}

\begin{document}
%\ninept
%
\maketitle
\begin{abstract}
Transcribed datasets typically contain speaker identity for each instance in the data. We investigate two ways to incorporate this information during training: \emph{Multi-Task Learning} and \emph{Adversarial Learning}. In multi-task learning, the goal is speaker prediction; we expect a performance improvement with this joint training if the two tasks of speech recognition and speaker recognition share a common set of underlying features. In contrast, adversarial learning is a means to learn representations invariant to the speaker. We then expect better performance if this learnt invariance helps generalizing to new speakers. While the two approaches seem natural in the context of speech recognition, they are incompatible because they correspond to opposite gradients back-propagated to the model. In order to better understand the effect of these approaches in terms of error rates, we compare both strategies in controlled settings. Moreover, we explore the use of additional un-transcribed data in a semi-supervised, adversarial learning manner to improve error rates. Our results show that deep models trained on big datasets already develop invariant representations to speakers without any auxiliary loss. When considering adversarial learning and multi-task learning, the impact on the acoustic model seems minor. However, models trained in a semi-supervised manner can improve error-rates.

\end{abstract}

\input{01_intro}

\input{03_advmulti}

\input{04_models}
\input{05_exp}

%\input{06_futurework}
\input{06_conclusion}

\begin{small}
\bibliographystyle{IEEEbib}
\bibliography{mybib}
\end{small}

\end{document}

%% file: definitions.tex
\usepackage{bm}
\usepackage{amssymb,amsmath,graphicx,bbm,color,booktabs}
\usepackage{amsopn}

\newcommand{\vr}{\bm{r}}               
\newcommand{\vs}{\bm{s}}

\newcommand{\vx}{\bm{x}}

\newcommand{\mr}{\bm{R}}

\newcommand{\mx}{\bm{X}}          
\newcommand{\my}{\bm{Y}}

    \newcommand{\Lc}{\mathcal{L}}

\renewcommand{\S}{\mathcal{S}}

\renewcommand{\eqref}[1]{Eq.~(\ref{#1})}

\def \y{{\mathbf y}}

\def \L{{\mathcal L}}

%% file: 01_intro.tex
% !TEX root =  main.tex
\section{Introduction}

The minimal components of a speech recognition dataset are audio recordings and their corresponding transcription. They also typically contain an additional information, which is the anonymous identifier of the speaker corresponding to each utterance. As speaker variations are one of the most challenging aspects of speech processing, this information can be leveraged to improve the performance and robustness of speech recognition systems.

Traditionally, the identity of the speakers has been used to extract speaker representations and use it as an additional input to the acoustic model~\cite{ivector_input1, ivector_input2}. This approach requires an additional feature extraction process of the input signal. Recently, two approaches have been proposed to leverage speaker information as another supervision to the acoustic model; Multi-Task Learning (MT)~\cite{caruana1998multitask,ronan_language} and Adversarial Learning (AL)~\cite{domain_adversarial_nn}. % \cite{mtl_speaker, tang2016multi, adversarial_speaker, adversarial_speaker_2, meng2018speaker}.
In the context of speech recognition, a way of performing MT or AL is to add a speaker classification branch in parallel of the main branch trained for transcription. All layers below the fork of the two branches receive gradients from both transcription loss and speaker loss.

In MT our goal is to jointly transcribe the speech signal together with classifying the speaker identity. Previous work has explored using auxiliary tasks such as gender or context \cite{mtl_overview}, as well as speaker classification \cite{mtl_speaker, tang2016multi}. For example, the authors in~\cite{mtl_speaker} propose to classify the speaker identity in addition to estimating the phoneme-state posterior probabilities used for ASR, they presented a non-negligible improvement in terms of \emph{Phoneme Error Rate} following the above approach.

An opposite approach is to assume that good acoustic representations should be invariant to any task that is not the speech recognition task, in particular the speaker characteristics. A method to learn such invariances is AL: the branch of the speaker classification task is trained to reduce its classification error, however the main branch is trained to maximize the loss of this speaker classifier. By doing so, it learns invariance to speaker characteristics. This approach has been previously used for speech recognition to learn invariance to noise conditions \cite{adversarial_noise}, speaker identity \cite{adversarial_speaker_2, meng2018speaker, adversarial_speaker} and accent~\cite{sun2018domain}. The authors in~\cite{meng2018speaker}, proposed to minimize the senone classification loss, and simultaneously maximize the speaker classification loss. This approach achieves a significant improvement in terms of \emph{Word Error Rate}. 

Our study arises from two observations. First, both the MT and AL approaches described above have been used with the same purpose despite being fundamentally opposed, and this raises the question of which one is the most appropriate choice to improve the performance of speech recognition systems. Secondly, they only differ by a simple computational step which consists in reversing the gradient at the fork between the speaker branch and the main branch. This allows for controlled experiments where both approaches can be compared with equivalent architectures and number of parameters.

In this work we focus on letter based acoustic models. In this setting, we are given audio recordings and their transcripts, and train a neural network to output letters with a sequence-based criterion. During training, we also have access to the identity of the speaker of each utterance. We perform a systematic comparison between MT and AL for the task of large vocabulary speech recognition on the Wall Street Journal dataset (WSJ) \cite{paul1992design}. Knowing whether the speaker information should be used to impact low level or high level representations is also a valuable information, so for each approach we experiment with three levels to fork the speaker branch to: lower layer, middle and upper layer. 

\textbf{Our Contribution:} 
(i) We observed that deep models trained on big datasets, already develop invariant representations to speakers with neither AL nor MT;
(ii) Both AL and MT do not have a clear impact on the acoustic model with respect to error rates; and
(iii) Using additional, un-transcribed speaker labeled data, (i.e. the speaker is known, but without the transcription), seems promising. It improves Letter-Error Rates (LER), even though these improvements did not transfer into Word-Error Rates (WER) improvements.

%\nico{It sounds like "we give you results, we will tell you how to interpret them in our next paper"}

%% file: 03_advmulti.tex
% !TEX root =  main.tex
\section{Adversarial vs. Multitask}
%In this section we formally describe AT and present the connection to MT in context of speaker classification for Automatic Speech Recognition systems.
Given a training set of $n$ transcribed acoustic utterances coupled with their speaker identity,  our goal is to utilize these speaker labels to improve transcription loss. 
  
Formally, let $\S = {(\mx_i, \my_i, s_i)}_{i=0}^n$ be the training set, in which each example is composed of three elements: a sequence of $m$ acoustic features $\mx_i = \{\vx_1, \cdots, \vx_m\}$ where each $x_j$ is a $d$-dimensional vector, a sequence of $k$ characters $\my_i = \{\y_1, \cdots, \y_k\}$ which are not aligned with $\mx_i$,  and a speaker label $s_i$. In the following subsections we present two common approaches to add the speaker identity as another supervision to the network.

\subsection{Adversarial Learning}
As speech recognition tasks should not depend on a specific set of speakers, in AL, we would like to learn an acoustic representation which is \emph{speaker-invariant} and at the same time \emph{characters-discriminative}. For that purpose, we consider the first $k$ layers of the network as an encoder, denoted by $E_r$ with parameters $\theta_r$, which maps the acoustic features $\mx_i$ to $\mr_i=\{\vr_1, \cdots, \vr_l\}$ where $l$ can be smaller than $m$. Then, the representations $\mr_i$ are fed into a decoder network $D_y$ with parameters $\theta_y$, to output the characters posteriors $p(\my_i|\mx_i;\theta_r, \theta_y)$. 
%as follows: 
% \begin{equation}
% D_y(\mr_i) =  D_y(E_r(\mx_i)) = p(\my_i | \mx_i; \theta_r, \theta_y)
% \end{equation}

Furthermore, we introduce another decoder network to classify the speaker, denote $D_s$. The above decoder, maps $\mr_i$ to the speaker posteriors $p(s_i|\mx_i;\theta_r, \theta_y)$.
% as follows: 
% \begin{equation}
% D_s(\mr_i) =  D_s(E_r(\mx_i)) = p(s_i | \mx_i; \theta_r, \theta_s)
% \end{equation}

In order to make the representation $\mr$ \emph{speaker-invariant}, we train $E_r$ and $D_s$ jointly using adversarial loss, where we optimize $\theta_r$ to \emph{maximize} speaker classification loss, and at the same time optimizing $\theta_s$ to \emph{minimize} the speaker classification loss.  Recall, we would like to make the representation $\mr$ \emph{characters-discriminative}, hence, we further optimize $\theta_y$ and $\theta_r$ to minimize the transcription loss. Therefore, the total loss is constructed as follows, 
\begin{equation}
\Lc(\theta_r, \theta_y, \theta_s) = \Lc_{acoustic}(\theta_r, \theta_y) - \lambda\Lc_{spk}(\theta_r, \theta_s)
\end{equation}
where $\Lc_{acoustic}(\theta_r, \theta_y)$ is the transcription loss, $\Lc_{spk}(\theta_r, \theta_s)$ is the speaker loss, and $\lambda$ is a trade-off parameter which controls the balance between the two. 
%Overall we would like to find the parameters $\hat{\theta}_y, \hat{\theta}_r,$ and $\hat{\theta}_s$ such that, 
%
%\begin{align*}
%\{\hat{\theta}_y, \hat{\theta}_r\} &= \min_{\theta_y, \theta_r} \L(\theta_r, \theta_y, \hat{\theta}_s) \\
%\hat{\theta}_s &= \max_{\theta_s} \L(\hat{\theta}_r, \hat{\theta}_y, \theta_s) \\
%\end{align*}
This minimax game will enhance the discriminative properties of $D_s$ and $D_y$ while at the same time push $E_r$ towards generating speaker-invariant representation. 

As suggested in ~\cite{domain_adversarial_nn,meng2018speaker}, we optimize the parameters using Stochastic Gradient Descent where we reverse the gradients of ${\partial{\Lc_{spk}}} / {\partial{\theta_r}}$~\cite{domain_adversarial_nn}.

\subsection{Multi-Task Learning}
\begin{figure} [h!]
    \begin{center}
  \includegraphics[width=.8\linewidth]{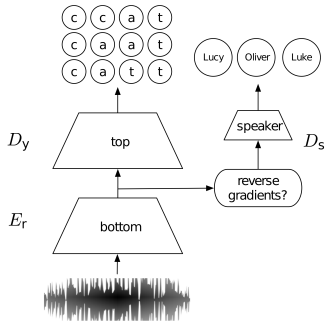}
    \end{center}
    \caption{An illustration of our architecture. We first feed the network a sequence of acoustic features, then we take the output of some intermediate representation an use it also to classify the speaker.}  \label{fig:model}
\end{figure}

Similarly to AL, in MT we consider an encoder $E_r$ with parameters $\theta_r$ and two decoders $D_y$ and $D_s$ with parameters $\theta_y$ and $\theta_s$ respectively. However, in MT, our goal is to minimize both $\Lc_{acoustic}(\theta_r, \theta_y)$ and $\Lc_{spk}(\theta_r, \theta_s)$. Therefore, the objective in MT case can be expressed as follows, 
\begin{equation}
\Lc(\theta_r, \theta_y, \theta_s) = \Lc_{acoustic}(\theta_r, \theta_y) + \lambda\cdot\Lc_{spk}(\theta_r, \theta_s)
\end{equation}
Practically, on forward propagation, both AL and MT act the same way. Yet, on back-propagation they act exactly the opposite. In AL, we reverse the gradients from $D_s$ before back-propagating into $E_r$. In contrast, in MT we sum together the gradients from $D_s$ and $D_y$. Figure~\ref{fig:model} depicts an illustration of the described model.

%% file: 04_models.tex
% !TEX root =  main.tex
\section{Models}

Our acoustic models are based on Gated Convolutional Neural Networks (Gated ConvNets)~\cite{dauphin2016language}, fed with 40 log-mel filterbank energies extracted every 10 ms with a 25 ms sliding window. Gated ConvNets stack 1D convolutions with Gated Linear Units (GLUs). 1D ConvNets were introduced early in the speech community, and are also referred as Time-Delay Neural Networks~\cite{peddinti2015time}. 

%Convolutional Neural Networks (ConvNets) \cite{lecun1995convolutional} with Gated Linear Units (GLUs) \cite{dauphin2016language}, fed with 40 log-mel filterbank energies extracted every 10 ms with a 25 ms sliding window. 
%
%The architecture is based on 1D Gated ConvNet. 1D ConvNets were introduced early in the speech community, and are also referred as Time-Delay Neural Networks (TDNNs) \cite{peddinti2015time}. Gated ConvNets \cite{dauphin2016language} stack 1D convolutions with Gated Linear Units (GLUs). 
We set $\L_{acoustic}(\theta_r, \theta_y)$ to be the AutoSeg Criterion (ASG) criterion \cite{collobert2016wav2letter,liptchinsky2017based}, which is similar to the Connectionist Temporal Classification (CTC) criterion \cite{graves2006connectionist}. 

Although speech utterances contain time dependent transcriptions, the speaker labels should be applied to the whole sequence. Hence, when optimizing for multi-task or adversarial learning we add a speaker branch which aggregates over all the time frames to classify the speaker. This pooling operation aggregates the time-dependent representations into a single sequence-level representation $\vs = g(\cdot)$. Two straightforward aggregation methods are to take the sum over all frames: $g(\cdot) = \sum_t \vr_t$ where $\vr_t$ is the representation at time frame $t$, or the max $g(\cdot) = \max_t (\vr_t)$, where the max is taken over each dimension. We propose to use a trade-off solution between these two methods, which is the LogSumExp \cite{palaz2016jointly}, $\vs = \frac{1}{\tau} \log \big ( \frac{1}{T} \sum_t e^{\tau \cdot \vr_t} \big )$, where $\tau$ is a hyper-parameter controls the trade-off between the sum and the max; with high values of $\tau$ the LogSumExp is similar to the max, while it tends to a sum as $\tau\rightarrow 0$. In practice, we used $\tau=1$.

We set $L_{speaker}(\theta_r, \theta_s)$ to be the Negative Log Likelihood (NLL) loss function. The overall network, together with $D_y$, is trained by back-propagation. At inference time, $D_s$, the speaker classification branch is discarded.

%% file: 05_exp.tex
\section{Experimental Results}
%\begin{table}[t!]
%\centering
%\caption{WSJ statistics. All timings besides total speaking time are in minutes. We report total speaking time in hours.}
%\begin{tabular}{lcccccc}\toprule
% & Num. of speakers & Avg.(m) & Max.(m) & Min.(m) & Total (h) \\
%\midrule
% & 283 & 17.3 & 31.4 & 5.3 &  81.5\\ 
%\bottomrule
%\end{tabular}
%\label{tab:dbstats}
%\end{table}

In the following section we describe our experimental results. First, we provide all the technical details and setups in Subsection~\ref{sec:setups}. Then, we describe our baselines and analyze their ability to encode information about the speaker in Subsection~\ref{sec:base}. Lastly, on Subsection~\ref{sec:wsj} we present our results on the WSJ dataset \cite{paul1992design}.

\subsection{Setups}
\label{sec:setups}

All models were composed of 17 gated convolutional layers followed by a weight normalization \cite{salimans2016weight}. We used a dropout layer after every GLU layer with dropout rate of 0.25. We optimized the models using SGD with two different learning rate values, we used learning rate of 1.4 for $E_r$ and $D_y$ and learning rate of 0.1 for $D_s$. We did not use momentum or weight decay. All the models were trained using WSJ dataset which contains 283 speakers with ~81.5 hours of speaking time.

We computed WER using our own one-pass decoder, which performs a simple beam-search with beam thresholding, histogram pruning and language model smearing \cite{steinbiss1994improvements}. We did not implement any sort of model adaptation before decoding, nor any word graph rescoring. Our decoder relies on KenLM \cite{heafield2013scalable} for the language modeling part, using a 4-gram LM trained on the standard data of WSJ \cite{paul1992design}. For LER, we used Viterbi decoding as in \cite{collobert2016wav2letter}.

The speaker branch, $D_s$, was composed of one gated convolutional layer of width 5 and 200 feature maps with weight normalization, followed by a linear layer. For MT models we used a static $\lambda$ value of 0.5. For AL models, we slowly increased the $\lambda$ from 0 to 0.2 using the following update scheme, $\lambda_i = 2 / (1 + e^{- p_i}) - 1$, where $p_i$ is a scaled version of the epoch number. The above technique was successfully explored in previous studies \cite{lample2017fader, domain_adversarial_nn}. In practice, we observed that reversing the gradients from the beginning of training can cause the network to diverge. To avoid that, we first trained the models without back-propagating the gradients from $D_s$ to $E_r$, (i.e. $\lambda = 0$). Then, we trained only $D_s$, and finally, we trained both of them jointly. We found the above procedure crucial for AL, however in MT the performance is equivalent to standard training. Thus, for comparison, we followed the above approach in both settings.

%Last, we define two controlled settings where we sampled subsets of the dataset in order to study the effect of the number of different speakers in the training set. In the both subsets, we sample $\sim$25 hours of speaking. In the first one, \maxspk, we forced the dataset to be balanced, i.e. every speaker has roughly the same amount of training data. In the second one, \minspk, we aim at using the minimum amount of speakers in order to reach the same number total time of speech. A detailed description of these datasets is given in Section~\ref{sec:abl}

\subsection{Representation Analysis}
\label{sec:base}

We would like to assess \emph{to what extent the network encodes the speaker identity?} The motivation for answering this question is two-fold (i) to analyze the networks' behavior with respect to the speaker information; and (ii) having a measurement of how much speaker information is encoded in the representation; this can provide an intuitive insight into how much error rates can be improved using speaker-adversarial learning.

To tackle this question, we follow a similar approach to the one proposed in \cite{adi2016fine, adi2017analysis}. We first trained a baseline network without the speaker classification branch, then we used it to extract representations of the speech signal from different layers in the network. Finally, we use these representations only, to train a model to classify the speaker identity. After training, we measure the model's accuracy to analyze the presence of speaker identity in the representations.

The basic premise of this approach is that if we cannot optimze a classifier to predict speaker identity based on representation from the model, then this property is not encoded in the representation, or rather, not encoded in an effective way, considered the way it will be used. Notice that we are not interested in improving speaker classification but rather to have a comparison between the models based on the classifier's classification accuracy.

For that purpose, we defined three settings, where we add the speaker branch after: (i) two convolutional layers, denote as $\inlayer$, (ii) eight convolutional layers, denote as $\midlayer$, and (iii) 15 convolutional layers, denote as $\outlayer$. For each setting we trained a classification model for 10 epochs, using the same architecture as the speaker classification branch.

\begin{figure} [t!]
  \begin{center}
  \includegraphics[width=0.75\linewidth]{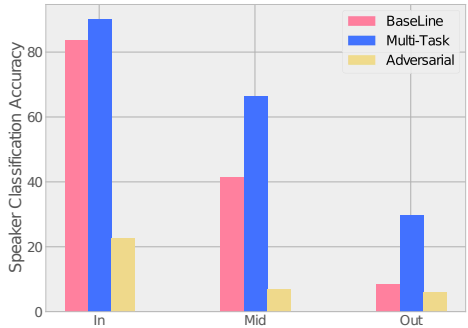}
  \end{center}
  \caption{Speaker classification accuracy using the representation from different layers of the network. Results are reported for Baseline, MT and AL models using three different settings, after the second, eighth, and fifteenth layers.}
  \label{fig:spkclass}
\end{figure}

Figure~\ref{fig:spkclass} presents the speaker classification accuracy for the baseline together with MT and AL using representations from $\inlayer, \midlayer$ and $\outlayer$. Notice, we observed a similar behavior in AL, MT, and the baseline, where the classification gets harder when extracting representations from deeper layers in the network. This implies that the network already develops speaker invariant representations during training. Even though, adding additional speaker loss can push it further and improve speaker invariance.

%\neil{we need to conclude by stating that even though the network already learns some speaker invariance by itself, an additional loss pushes it further}

%Those results provide us with more information on how much we can improve using adversarial training or multi-task training techniques.

%\vspace{-0.1cm}
\subsection{WSJ Experiments}
\label{sec:wsj}
Following our experiments in subsection~\ref{sec:base}, we trained MT and AL in three versions, $\inlayer, \midlayer$ and $\outlayer$. Table~\ref{tab:results} summarizes the results of our models, the baseline, and state-of-the-art systems.

\begin{table}[t!]
\caption{Letter Error Rates and Word Error Rates on the WSJ dataset.}
\label{tab:results}
    \centering
\scalebox{0.85}{\begin{tabular}{ccccccc}
\toprule
\multicolumn{2}{c}{Model} & \multicolumn{2}{c}{Nov93 dev} & \multicolumn{2}{c}{Nov92 eval} \\
 Type & Layer & LER & WER & LER & WER \\
\midrule
\multicolumn{2}{l}{\textit{Human} \cite{amodei2015deep}} & - & 8.1 & - & 5.0 \\
\multicolumn{2}{l}{BLSTMs \cite{chan2015deep}} & - & 6.6 & - & \textbf{3.5} \\
\multicolumn{2}{p{4.0cm}}{convnet BLSTM w/ additional data \cite{amodei2015deep}} & - & 4.4 & - & 3.6 \\
%\multicolumn{2}{p{2.7cm}}{convnet BGRU w/ data augm. \cite{zhou2017improving}} & - & - & - & 9.2 & - & 5.5 \\
%\multicolumn{2}{l}{Chan et al. 2015 \cite{chan2015deep}} & - & - & - & 6.6 & - & 3.5 \\
%\multicolumn{2}{l}{Amodei et al. 2015 \cite{amodei2015deep}} & - & - & - & 4.4 & - & 3.1 \\
%\multicolumn{2}{l}{Zhou et al. 2018 \cite{zhou2018policy}} & - & - & - & 9.2 & - & 5.5 \\
\multicolumn{2}{l}{Our Baseline} & 7.2 & \textbf{9.7} & 4.9& 5.9\\
  \midrule
  \multirow{3}{*}{\rotatebox[origin=c]{90}{Mult.}} 
 & \inlayer & 7.2 & 9.9 & 4.8 & 5.8\\
 & \midlayer & 7.3 & 9.8 & 4.9 & 5.9\\
 & \outlayer & 7.3 & 10.1 & 4.8 & 5.7\\
  \midrule
\multirow{3}{*}{\rotatebox[origin=c]{90}{Adv.}} 
 & \inlayer & 7.2 & 9.9 & 4.7 & 5.7\\
 & \midlayer & 7.3 & 10.0 & 4.8 & 5.8\\
 & \outlayer & 7.2 & 9.8 & 4.7 & \textbf{5.6}\\
  \midrule
\multirow{3}{*}{\rotatebox[origin=c]{90}{Semi.}} 
 & \inlayer & \textbf{6.8} & 9.8 & \textbf{4.6} & 6.2\\
 & \midlayer & 6.8 & \textbf{9.7} & 4.7 & 6.0\\
 & \outlayer & 6.9 & 10.0 & 4.6 & 6.3\\
\bottomrule
\end{tabular}}
\end{table}

Using both AL and MT seems to have a minor effect on transcription modeling when training on WSJ dataset. Both approaches achieve equal or somewhat worse LER than the baseline on the development set, and improve over the test set, with AL performs slightly better than MT. When considering WER, both AL and MT improve results over the baseline on the test set, however none of them improve over the development set. These results are somewhat counter-intuitive since both approaches attain similar results, however can be considered as opposite paradigms. One explanation is that both methods act as another regularization term. In other words, although the two approaches push the gradients towards opposite directions, when treated carefully, they both impose another constraint on the models' parameters~\cite{ruder2017overview, nski2017adversarial, cohen2018cross}. 

\subsection{Adversarial Semi-Supervised Learning}
Lastly, we wanted to investigate whether the observed improvements were limited by the relatively small amount of speakers (283) in the train set of WSJ.  For that purpose, we ran a semi-supervised experiment were we control the amount of transcribed data by using only speaker labeled examples. We add $\approx$ 35 hours of speech using another 132 speakers, from the Librispeech corpus~\cite{panayotov2015librispeech}, and train the network in an adversarial fashion. For the new training examples we do not use the transcriptions, but only the speaker identity, hence optimizing the speaker loss only. Results are summarized on Table~\ref{tab:results}. Using additional speaker labeled data, seems beneficial in terms of LER: it improves over the baseline and previous AL results in all settings. 

%Notice, in the current setting we get a significant LER improvement both on the test and development sets, this may imply a more generalized representation. 

%\nico{not sure what that is supposed to mean... the reader hopes that what you call "improve" means "improve both on dev and test". What is a "generalized representation"?} 

On the other hand, when considering WER, none of the settings improve over the baseline. This misalignment between the LER and WER results can be either because we constrained the model's outputs by LM at decoding time, hence WER results are greatly affected by the type of LM used, or because the LER improvements are on meaningless parts in the sequence. The analysis of this observations as well as jointly optimizing acoustic model and LM are left for future research. 
%\nico{I like the idea of the two possible explanations... But as a reader I would really like to say "it's not future work, it is very much work that should be done in this paper"} 

%Lastly, since AL performs slightly \neil{I think we should avoid this adverb} better than MT, we wanted to explore if the use of additional, partially labeled data (only speaker identity is provided) can further improve results. For that purpose we sampled additional $\sim$ 35 hours of speaking from Librispeech corpus~\cite{panayotov2015librispeech} using another \todo{100} speakers. 
%
%Also in this case, we suspect it is due to language model scaling. 
% 
% \neil{Again, I don't think we can avoid a more convincing explanation than just mentioning the scaling issue}

%we explored the use of additional, partially labeled data (only speaker identity is provided) in AL setting. 
%
%We experiment with AL only, due to the natural behavior of the network to develop invariance to speakers. 

%% file: 06_conclusion.tex
% !TEX root =  main.tex
%\vspace{-0.1cm}
\section{Conclusion and Future Work}
%We conducted a controlled comparison between speaker AL and speaker MT for automatic speech recognition on WSJ dataset. 
%We used three different settings for both methods, together with semi-supervised settings. In addition, we analyzed the amount of speaker information already embedded in the network. Results suggests that both methods can improve LER over the baseline, however when considering WER things are ambiguous. 

In this paper, we studied AL and MT in the context of speech recognition. We showed that deep models already learn speaker-invariant representations, but can still benefit from semi-supervised speaker adversarial learning.

%already learn speaker-invariant representations, and we obtained promising results in semi-supervised setups.

For future work, we would like to examine the effect of semi-supervised learning for low-resource languages.